\documentclass[10pt,twocolumn,letterpaper]{article}

\usepackage[pagenumbers]{cvpr} 

\usepackage{amsmath}

\usepackage{amsthm}
\newtheorem{definition}{Definition}
\newtheorem{theorem}{Theorem}
\newtheorem{remark}{Remark}

\usepackage{algorithm}
\usepackage{algpseudocode}
\usepackage{bm}
\usepackage{url}
\usepackage{float}
\usepackage{multirow}
\usepackage{longtable}
\usepackage{booktabs} 
\usepackage{subcaption}
\usepackage{caption}
\usepackage{array}

\definecolor{cvprblue}{rgb}{0.21,0.49,0.74}
\usepackage[pagebackref,breaklinks,colorlinks,allcolors=cvprblue]{hyperref}

\def\paperID{32393} 
\def\confName{CVPR}
\def\confYear{2026}

\title{Scaling Laws for Black-box Adversarial Attacks}

\author {
    Chuan Liu\textsuperscript{\rm 1,\rm 2,\rm 3}\footnotemark[1] \hspace{0.5cm}
    Huanran Chen\textsuperscript{\rm 1,\rm 3}\footnotemark[1] \hspace{0.5cm}
    Yichi Zhang \textsuperscript{\rm 1,\rm 3} \hspace{0.5cm}
    Jun Zhu \textsuperscript{\rm 1,\rm 3} \hspace{0.5cm}
    Yinpeng Dong \textsuperscript{\rm 1,\rm 3} \hspace{0.5cm} \\
     \textsuperscript{\rm 1} Dept. of Comp. Sci. and Tech., Institute for AI, Tsinghua-Bosch Joint ML Center, \\
    THBI Lab, BNRist Center, Tsinghua University, Beijing, 100084, China\\
    \textsuperscript{\rm 2} IIIS, Tsinghua University \hspace{1cm}
    \textsuperscript{\rm 3} RealAI\\
    {\tt\small \{liuchuan22, zyc22\}@mails.tsinghua.edu.cn, huanran\_chen@outlook.com} \\ {\tt\small \{dongyinpeng, dcszj\}@mail.tsinghua.edu.cn}
}

\begin{document}
\maketitle

\renewcommand{\thefootnote}{\fnsymbol{footnote}}
\footnotetext[1]{Equal contribution.}
\footnotetext[2]{Our code is available at \url{https://github.com/liuchuan22/ScaleUpAttack}.}

\begin{abstract}
Adversarial examples exhibit cross-model transferability, enabling threatening black-box attacks on commercial models. Model ensembling, which attacks multiple surrogate models, is a known strategy to improve this transferability. However, prior studies typically use small, fixed ensembles, which leaves open an intriguing question of whether scaling the number of surrogate models can further improve black-box attacks. In this work, we conduct the first large-scale empirical study of this question. We show that by resolving gradient conflict with advanced optimizers, we discover a robust and universal log-linear scaling law through both theoretical analysis and empirical evaluations: the Attack Success Rate (ASR) scales linearly with the logarithm of the ensemble size $T$. We rigorously verify this law across standard classifiers, SOTA defenses, and MLLMs, and find that scaling distills robust, semantic features of the target class. Consequently, we apply this fundamental insight to benchmark SOTA MLLMs. This reveals both the attack's devastating power and a clear robustness hierarchy: we achieve 80\%+ transfer attack success rate on proprietary models like GPT-4o, while also highlighting the exceptional resilience of Claude-3.5-Sonnet. Our findings urge a shift in focus for robustness evaluation: from designing intricate algorithms on small ensembles to understanding the principled and powerful threat of scaling.
\end{abstract}

    
\section{Introduction}
\label{sec:intro}

Over the past decade, deep learning has achieved remarkable advancements across various computer vision tasks, leading to its widespread applications \cite{lecun2015deep}. 
However, deep neural networks (DNNs) remain susceptible to adversarial examples \cite{goodfellow2014explaining, szegedy2013intriguing} -- perturbations that are subtle and nearly imperceptible but can significantly mislead the model, resulting in incorrect predictions. 
Adversarial attacks have garnered considerable interest due to their implications for understanding DNN functionality \cite{dong2017towards}, assessing model robustness \cite{zhang2024benchmarking}, and informing the development of more secure algorithms \cite{madry2017towards}.

\begin{figure}[t]
    \centering
    \includegraphics[width=0.99\columnwidth]{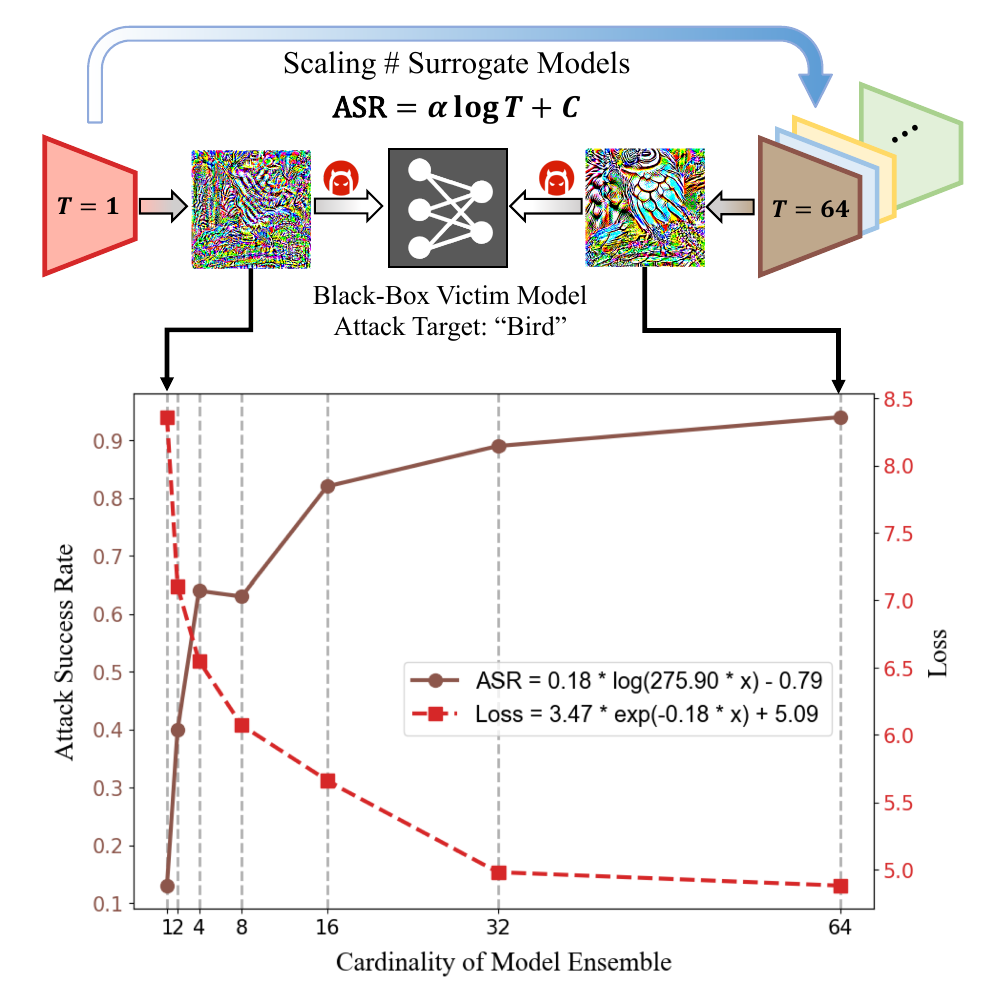}
    \caption{Key observation of scaling laws for the transfer-based black-box attack. By scaling the number of surrogate models in the ensemble, we observe that the attack success rate and the discriminative loss on the target model follow a log-linear scaling law. }
    \vspace{-2ex}
    \label{fig:Teaser}
\end{figure}

Adversarial attacks are typically categorized into white-box and black-box attacks based on whether the attacker has knowledge of the architecture and parameters of the target model. Given the greater risks with commercial services, e.g., GPT-4~\cite{openai2024gpt4technicalreport}, black-box attack methodologies~\cite{liu2016delving,dong2018boosting} have attracted growing attention. One prominent approach within black-box attacks is the transfer-based attack, which approximately optimizes the adversarial examples against the black-box model by minimizing the discriminative loss on a selection of predefined surrogate models.
Researchers have devised advanced optimization algorithms \cite{madry2017towards, wang2021enhancing, lin2019nesterov} and data augmentation techniques \cite{long2022frequency, dong2019evading} in transfer-based attacks.
Model ensembling \cite{liu2016delving,dong2018boosting} further facilitates these methods by leveraging multiple surrogate models simultaneously to improve the transferability of adversarial examples.
By averaging the gradients or predictions from different models, ensemble attack methods seek to increase the transferability across various target models \cite{dong2018boosting, chen2023rethinking}.

Despite the progress in ensemble attack algorithms, the number of surrogate models used in previous research is often limited~\cite{chen2023rethinking,zhang2024does,yao2025understandingmodelensembletransferable}. Recent advancement in large language models shows that scaling training data can significantly improve the generalization ability \cite{kaplan2020scalinglawsneurallanguage}. Since the transfer-based attack is also a generalization problem that expects adversarial examples from surrogate models to generalize to target models, while the training data are surrogate models rather than text or images~\cite{dong2018boosting,chen2023rethinking}, this equivalence prompts an intriguing question: \textit{Can we achieve significant improvements in transfer-based attacks by scaling the number of surrogate models, akin to the scaling of data and model parameters in large language models like GPT?}

In this work, we conduct the first large-scale empirical investigation into the scaling laws of black-box adversarial attacks, given that the gradient conflict issue is properly resolved (Sec.~\ref{sec:algorithms}). Our central finding is the discovery of a robust and universal log-linear scaling law: the Attack Success Rate (ASR) on black-box targets scales linearly with the logarithm of the ensemble cardinality $T$. We provide an asymptotic analysis on the convergence rate of generation error and adversarial examples under idealized IID assumptions (Sec.~\ref{subsec:3.1}). Our core contribution, however, is the empirical discovery and verification of the precise log-linear form of these scaling laws (Sec.~\ref{subsec:discovery}). We rigorously demonstrate the universality of these scaling laws across diverse target families (Sec.~\ref{subsec:verification}), and provide a qualitative explanation for this phenomenon (Sec.~\ref{subsec:6.1}): as $T$ increases, the optimizer is forced to discard model-specific, non-robust features and instead learns the robust, semantic features of the target class, resulting in highly transferable perturbations.

Inspired by this fundamental insight, we then extend our scaling laws to attack the foundational vision encoders (e..g., CLIP) of modern MLLMs (Sec.~\ref{sec:vision_encoder}) to demonstrate its practical power as both a SOTA attack and a novel robustness benchmark. We achieve devastatingly high ASR on top-tier models, including 90.0\% ASR on Qwen3-VL-235B and 85.0\% ASR on GPT-4o. Meanwhile, our scaling benchmark reveals a stark hierarchy in SOTA model robustness. We highlight the exceptional resilience of Claude-3.5-Sonnet in sharp contrast to the high vulnerability of Gemini-2.5-Pro. These findings show that scaling is a critical, measurable threat vector, and we urge the community to adopt large-scale robustness evaluation, rather than designing intricate algorithms on small ensembles, to build and validate genuinely robust and trustworthy foundation models.

\section{Backgrounds}
\label{sec:backgrounds}

In this section, we provide necessary backgrounds, including the problem formulations for different attack settings and the prerequisite algorithms.

\subsection{Problem Formulations}
\label{sec:problem_formulations}

Our work investigates scaling laws across multiple black-box attack scenarios. We formulate the settings below.

\subsubsection{Attacks on Image Classifiers}

Let $\mathcal{F}$ denote the set of all image classifiers. The vulnerability of such classifiers to adversarial examples has been a long-standing area of research \cite{szegedy2013intriguing, goodfellow2014explaining}. Given a natural image $\bm{x}_{\text{nat}}$ with its true label $\bm{y}_{\text{nat}}$, the goal of a transfer-based attack is to craft an adversarial example $\bm{x}$ within an $\ell_{\infty}$ perturbation budget $\epsilon$ that fools black-box target models.

\textbf{Targeted Attack.}
The primary focus of our work is the challenging targeted attack setting. Given a target class $\bm{y}$ (where $\bm{y} \neq \bm{y}_{\text{nat}}$), the goal is to craft an $\bm{x}$ that is misclassified as $\bm{y}$. This can be formulated as minimizing the expected loss over all models in $\mathcal{F}$:
\begin{equation}
  \min_{\bm{x}} \mathbb{E}_{f\in \mathcal{F}} \left[\mathcal{L} (f(\bm{x}), \bm{y})\right], \quad \text{s.t.} \left\lVert \bm{x} - \bm{x}_{\text{nat}}\right\rVert_{\infty} \leq \epsilon,
  \label{eq:constrained_optimization}
\end{equation}
where $\mathcal{L}$ is a loss function (e.g., cross-entropy). In practice, we approximate this by ensembling a limited set of $T$ surrogate classifiers $\{f_i\}_{i=1}^T \subset \mathcal{F}$:
\begin{equation}
    \min_{\bm{x}} \frac{1}{T} \sum_{i=1}^{T} \mathcal{L} (f_i(\bm{x}), \bm{y}), \quad \text{s.t.} \left\lVert \bm{x} - \bm{x}_{\text{nat}}\right\rVert_{\infty} \leq \epsilon.
    \label{eq:empirical_loss}
\end{equation}

\textbf{Untargeted Attack.}
We also validate our scaling laws in the untargeted setting in Sec.~\ref{sec:discussion:scaling_general}, where the goal is to mislead the classifier to \textbf{any} class other than $\bm{y}_{\text{nat}}$. The objective is to maximize the loss with respect to original $\bm{y}_{\text{nat}}$:
\begin{equation}
    \max_{\bm{x}} \frac{1}{T} \sum_{i=1}^{T} \mathcal{L} (f_i(\bm{x}), \bm{y}_{\text{nat}}), \quad \text{s.t.} \left\lVert \bm{x} - \bm{x}_{\text{nat}}\right\rVert_{\infty} \leq \epsilon.
    \label{eq:empirical_loss_untargeted}
\end{equation}

\subsubsection{Attacks on Vision Encoders}
\label{sec:clip_formulation}
Contrastive Language-Image Pretraining (CLIP) is a popular self-supervised framework that can pretrain large-scale language-vision models on web-scale text-image pairs via contrastive learning \cite{radford2021learningtransferablevisualmodels, palatucci2009zero, lampert2009learning}. Based on the generalization ability of CLIP models, previous works have utilized them as surrogate models in model ensembles \cite{dong2023robust, anonymous2024tuap, fort2024ensembleeverywheremultiscaleaggregation}. 

We extend our analysis to attack these vision encoders, which are foundational to modern MLLMs. Instead of classifying to a specific logit, the goal is to craft an image $\bm{x}$ whose embedding $f^I(\bm{x})$ is close to the text embedding $\bm{e}_{\text{tar}}$ of a target label $\bm{y}$. This target embedding is typically pre-computed using a text encoder, $f^T$. The objective function is adapted by replacing the classifier $f_i$ with a surrogate image encoder $f_i^I$ and the cross-entropy loss $\mathcal{L}$ with a similarity-based loss $\mathcal{L}_{\text{sim}}$ (e.g., cosine embedding loss):
\begin{equation}
    \min_{\bm{x}} \frac{1}{T} \sum_{i=1}^{T} \mathcal{L}_{\text{sim}} (f_i^I(\bm{x}), \bm{e}_{\text{tar}}), \quad \text{s.t.} \left\lVert \bm{x} - \bm{x}_{\text{nat}}\right\rVert_{\infty} \leq \epsilon,
    \label{eq:empirical_loss_clip}
\end{equation}
where $\bm{e}_{\text{tar}} = f^T(\bm{y})$ is the fixed target text embedding.

\subsection{Prerequisite Algorithms}
\label{sec:algorithms}

Researchers have developed sophisticated algorithms to enhance adversarial transferability, which can be categorized into gradient optimizers, input transformations, and ensemble strategies.

\subsubsection{Gradient Optimizers}
These methods refine the gradient update step itself. A pioneering example is \textbf{MI-FGSM}~\cite{dong2018boosting}, which stabilizes gradients by integrating a momentum term:
\begin{equation}
    \bm{g}_{t+1} = \mu \cdot \bm{g}_t + \frac{\nabla_{\bm{x}} \mathcal{L}(\bm{x_t}, \bm{y})}{\left\lVert \nabla_{\bm{x}} \mathcal{L}(\bm{x_t}, \bm{y}) \right\rVert_{1}},
    \label{eq:MI_FGSM_momentum}
\end{equation}
with update $\bm{x}_{t+1} = \bm{x}_t + \alpha \cdot \operatorname{sign} (\bm{g}_{t+1})$. Other methods like VMI-FGSM~\cite{wang2021enhancing} further refine this momentum concept.

\subsubsection{Input Transformations}
These methods apply data augmentation during the attack to find more generalizable adversarial patterns. Prominent examples include Diverse Inputs (DI)~\cite{xie2019improving} and Spectrum Simulation Attack (SSA)~\cite{long2022frequency}. SSA, a state-of-the-art method, applies spectrum transformations $\mathcal{T}$ and computes gradients on the transformed input $\mathcal{T}(\bm{x})$.

\begin{figure}[t]
    \centering
    \includegraphics[width=0.99\columnwidth, trim=1.8cm 17.2cm 2cm 3cm, clip]{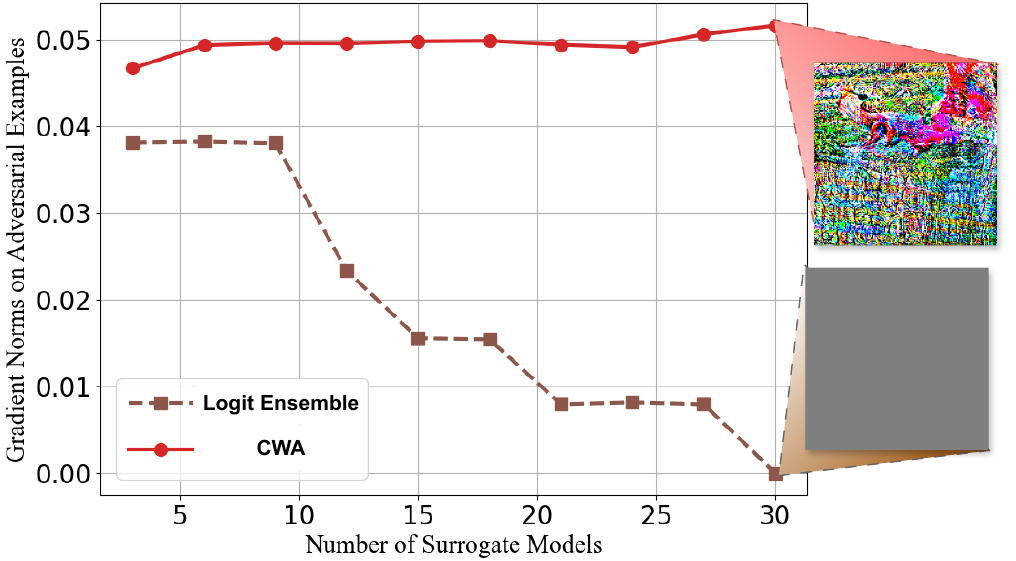}
    \caption{Comparisons between CWA and a na\"ive ensemble (MI-FGSM with logit averaging). As $T$ increases, the na\"ive method suffers optimization stagnation (gradient norm collapses) due to gradient conflict. Advanced methods like CWA, which align gradients, avoid this issue entirely.}
    \vspace{-2ex}
    \label{fig:CWA_FGSM}
\end{figure}

\subsubsection{Ensemble Strategies and Gradient Conflict}
The above components can be applied to ensemble attacks. The most straightforward method, often used with MI-FGSM, is logit ensembling~\cite{dong2018boosting}, where the optimizer computes gradients over the sum of all models' logits.

While simple, this strategy exposes a critical flaw: \textbf{gradient conflict}. We discover that as the number of surrogates ($T$) increases, the gradients from different models, which often lack semantics and have random directions, tend to cancel each other out. As shown in Fig.~\ref{fig:CWA_FGSM}, this conflict causes the ensemble gradient norm to collapse, leading to complete optimization stagnation.

To address this, advanced optimizers are required. These methods aim to find a common direction of vulnerability rather than simply averaging gradients. They typically work by reducing gradient variance, such as SVRE~\cite{xiong2022stochastic}, or by encouraging gradient alignment, which is the approach taken by the Common Weakness Attack (CWA)~\cite{chen2023rethinking}. CWA seeks common vulnerabilities by encouraging flatter loss landscapes and minimizing the second-order Taylor series:
\begin{equation}
        \frac{1}{T}\sum_{i=1}^T \left[ \mathcal{L}(f_i(\bm{p}_i, \bm{y}) + \frac{1}{2} (\bm{x} - \bm{p}_i)^\top H_i (\bm{x} - \bm{p}_i) \right],
        \label{eq:CWA_objective}
\end{equation}
where $\bm{p}_i$ is the closest optimum of model $f_i$ to $\bm{x}$ and $H_i$ is the Hessian. As Fig.~\ref{fig:CWA_FGSM} demonstrates, CWA effectively avoids the optimization stagnation seen in MI-FGSM.

\textbf{Our Chosen Algorithm: SSA-CWA.}
For our main experiments, we require an algorithm that both resolves the gradient conflict and achieves state-of-the-art performance. We therefore select SSA-CWA, which combines the advanced input transformation of SSA~\cite{long2022frequency} with the robust gradient optimization of CWA~\cite{chen2023rethinking} (pseudocode provided in Appendix~\ref{app:ssa_cwa}). While leveraging SSA-CWA as our core algorithm, we also explore the scaling laws of other optimizers and validate our gradient conflict hypothesis in Sec.~\ref{sec:discussion:scaling_general}.

\section{Scaling Laws for Adversarial Attacks}
\label{sec:scaling}

In this section, we present our key findings on scaling laws for black-box adversarial attacks. Sec.~\ref{subsec:3.1} introduces the theoretical motivation for our study. Sec.~\ref{subsec:setup} details our experimental setup for generating adversarial examples. Sec.~\ref{subsec:discovery} presents the initial discovery of the empirical scaling laws on standard classifiers. Finally, \cref{subsec:verification} verifies the universality of these laws by testing them against state-of-the-art defended models and across diverse MLLMs.

\subsection{Motivation}
\label{subsec:3.1}

First, we observe that in most practical scenarios, $\bm{x}_{\text{nat}}$ and $\bm{y}$ are predefined and fixed. Consequently, Eq.~\eqref{eq:constrained_optimization} and Eq.~\eqref{eq:empirical_loss} can be considered as functions of variable $\bm{x}$ only.

\begin{definition}
\label{def:minimizer}
    Define $\bm{x}^*$ and $\hat{\bm{x}}$ to be the minimizers of Eq.~\eqref{eq:constrained_optimization} and Eq.~\eqref{eq:empirical_loss}, respectively, which can be written as
    \begin{equation}
        \begin{aligned}
        &\bm{x}^* = \arg\min_x \mathbb{E}_{f \in \mathcal{F}} [\mathcal{L} (f(\bm{x}), \bm{y})],\\
        &\hat{\bm{x}} = \arg \min _x \frac{1}{T} \sum_{i=1}^{T} \mathcal{L} (f_i(\bm{x}), \bm{y}), 
        \end{aligned}
    \end{equation}
    with the same $\ell_\infty$ norm constraints given in Eq.~\eqref{eq:constrained_optimization} and ~\eqref{eq:empirical_loss}.
\end{definition}

\begin{definition}
\label{def:losses}
    Define $\mathbb{L}(\cdot)$ as the population loss in Eq.~\eqref{eq:constrained_optimization}, and $\mathbb{L}(f_i(\cdot))$ as the loss on one surrogate model $f_i$: 
    \begin{equation}
        \begin{aligned}
            &\mathbb{L}(\bm{x}) = \mathbb{E}_{f \in \mathcal{F}} [\mathcal{L} (f(\bm{x}), \bm{y})], \\
            &\mathbb{L}(f_i(\bm{x})) = \mathcal{L} (f_i(\bm{x}), \bm{y}).
        \end{aligned}
    \end{equation}
\end{definition}

These definitions are used to simplify our notations and better demonstrate our proofs. Then, we bring our key observations that are shown in the following theorem.

\begin{theorem}
\label{thm:asymptotic_general}
Suppose that model ensemble \(\{f_i\}_{i=1}^T\) is i.i.d. sampled from some distribution on \(\mathcal{F}\), and \(\hat{\bm{x}}\) goes to \(\bm{x}^*\) as \(T \to \infty\). Denote \(\overset{d}{\to}\) as convergence in distribution, and $\mathrm{Cov}$ as the covariance matrix. Then we have the following asymptotic bound as \(T \to \infty\):
\begin{equation}
    \begin{aligned}
        &\sqrt{T}(\hat{\bm{x}}-\bm{x}^*) \overset{d}{\to}
    \mathcal{N}(0, (\nabla^2 \mathbb{L}(\bm{x}^*))^{-1}
    \mathrm{Cov}(\mathbb{L}(f_i))
    (\nabla^2 \mathbb{L}(\bm{x}^*))^{-1}
    ), \\
    &T (\mathbb{L}(\hat{\bm{x}}) - \mathbb{L}(\bm{x}^*) )\overset{d}{\to} \frac{1}{2}\|S\|_2^2  , \text{ with} \\
    & S \sim \mathcal{N}(0, (\nabla^2 \mathbb{L}(\bm{x}^*))^{1/2}
    \mathrm{Cov}(\mathbb{L}(f_i))
    (\nabla^2 \mathbb{L}(\bm{x}^*))^{1/2}).
    \end{aligned}
\end{equation}
\end{theorem}

Proof of Theorem \ref{thm:asymptotic_general} can be found in Appendix \ref{app:proof_thm1}.

\begin{remark}
    Theorem \ref{thm:asymptotic_general} suggests that the adversarial examples converge in $\mathcal{O}(\frac{1}{\sqrt{T}})$, while the difference between population loss and empirical loss converges in $\mathcal{O}(\frac{1}{T})$.
\end{remark}

When we replace $\mathcal{L}$ with the cross-entropy loss in Theorem~\ref{thm:asymptotic_general}, we obtain Theorem~\ref{thm:asym_negative_likelyhood} by some simple derivations.

\begin{theorem}
\label{thm:asym_negative_likelyhood}
    Following the notation and assumptions in \cref{thm:asymptotic_general}, when using the cross-entropy loss, we have:
    \begin{equation}
    \begin{aligned}
        &\sqrt{T}(\hat{\bm{x}}-\bm{x}^*) \overset{d}{\to}
    \mathcal{N}(0, (\nabla^2 \mathbb{L}(\bm{x}^*))^{-1}), \\
    &T (\mathbb{L}(\hat{\bm{x}}) -\mathbb{L}(\bm{x}^*) )\overset{d}{\to} \frac{1}{2}\|\mathcal{N}(0, \bm{I})\|_2^2.
    \end{aligned}
    \end{equation}
\end{theorem}

\begin{remark}
    Theorem~\ref{thm:asym_negative_likelyhood} holds since \(\mathrm{Cov}(\mathbb{L}(f_i(\bm{x}^*))) = \nabla^2 \mathbb{L}(\bm{x}^*)\) for the cross-entropy loss. Therefore, we have \(2T(\mathbb{L}(\hat{\bm{x}}) - \mathbb{L}(\bm{x}^*) ) \overset{d}{\to} \chi^2(D)\) where \(D\) is the dimension of the input, which means \(\mathbb{L}(\hat{\bm{x}}) -\mathbb{L}(\bm{x}^*)\) converges in \(\mathcal{O}(\frac{D}{2T})\).
\end{remark}

These theorems provide the core theoretical motivation for our work: increasing the ensemble cardinality $T$ is a principled approach, as it provably closes the gap between the empirical loss and the true population loss. However, we must acknowledge a crucial gap between theory and practical setting of black-box attacks. The theorems rely on i.i.d. assumptions that are not met in practice but for a theoretical convenience. In practice, our surrogate models are typically pre-trained on similar datasets (e.g., ImageNet) and share strong architectural biases, making them highly correlated~\cite{ilyas2019adversarialexamplesbugsfeatures}. Consequently, the asymptotic $\mathcal{O}(1/T)$ convergence rate derived in Remark \ref{thm:asym_negative_likelyhood} is unlikely to hold.

Therefore, we posit that while the theory provides a strong \textit{qualitative} motivation (i.e., scaling $T$ should improve transferability), it does not offer a \textit{quantitative prediction} for the \textit{empirical} scaling law. The central question of how ASR and loss actually behave as $T$ increases in a practical setting remains open. This motivates our large-scale empirical investigation in the following sections to discover the true scaling laws governing black-box attacks.

\subsection{Experimental Setup}
\label{subsec:setup}

\begin{figure*}[ht]
    \centering
    \includegraphics[width=0.99\textwidth, trim=0cm 0cm 0cm 0cm, clip]{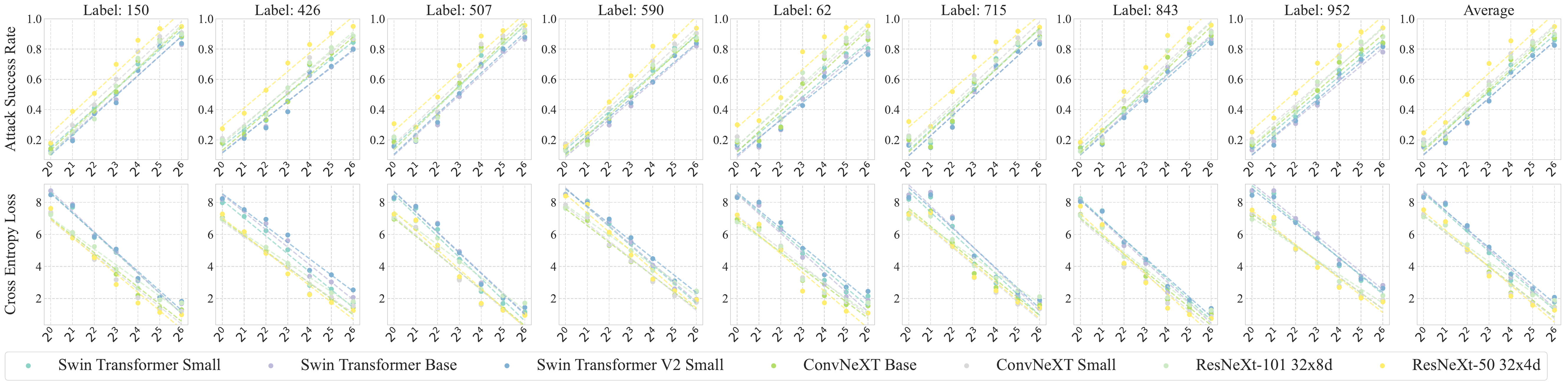}
    \caption{Scaling Laws over model ensemble in a black-box setting. We plot the x-axis on a base-2 logarithmic scale, and we measure both the attack success rate (ASR) and average cross entropy-loss of target models. The cardinality of the model ensembles is varied from $2^0$ to $2^6$, with models randomly selected for each ensemble. We fit the results of selected target models by lines for better demonstration.}
    \vspace{-3ex}
    \label{fig:scaling_laws}
\end{figure*}

To investigate the relationship between the number of surrogate models and attack efficacy, we first define our comprehensive experimental setup. This setup is used to generate a single, fixed set of adversarial examples, which are then evaluated across diverse target models to discover and verify our proposed scaling laws.

\subsubsection{Adversarial Example Generation}

\textbf{Surrogate Pool.}
To ensure the transferability and diversity of our attack, we select a large pool of 64 models from Torchvision \cite{torchvision2016} and Timm \cite{rw2019timm}. This pool includes a wide variety of architectures, hyperparameters, and training datasets, such as AlexNet \cite{krizhevsky2012imagenet}, DenseNet \cite{huang2017densely}, EfficientNet \cite{tan2019efficientnet}, ResNet \cite{he2016deep}, ViT \cite{dosovitskiy2020image}, DeiT \cite{touvron2021training}, and many others. The full list is available in Appendix \ref{app:settings_scaling_laws}.

\textbf{Ensemble Sampling and Attack Algorithm.}
We generate adversarial examples by attacking ensembles of surrogate models. For each attack, an ensemble is constructed by randomly sampling $T$ models from our 64-model pool. We vary the ensemble cardinality $T$ across a logarithmic scale from $2^0$ to $2^6$. 
We use the SSA-CWA attack \cite{chen2023rethinking} for 40 iteration steps, with a perturbation budget of $\epsilon = 8 / 255$ under the $\ell_\infty$ norm. The loss function $\mathcal{L}$ is the standard cross-entropy loss. This process yields 7 distinct sets of adversarial examples, one for each value of $T$.

\subsubsection{Target Model Families}
\label{subsubsec:target_families}

To test the universality of any observed scaling phenomena, we evaluate our pre-generated adversarial examples against three distinct families of black-box target models:

\begin{itemize}
    \item \textbf{Family 1: Standard Image Classifiers.} We select 7 widely-used models, including Swin-Transformers \cite{liu2021swin}, ConvNeXt \cite{liu2022convnet}, and ResNeXt \cite{xie2017aggregated}.
    
    \item \textbf{Family 2: Defended Models.} We select 6 models representing SOTA defenses, including adversarially trained models (FGSMAT \cite{kurakin2016adversarial}, EnsAT \cite{tramer2017ensemble}) and purification defenses (HGD \cite{liao2018defenseadversarialattacksusing}, BDR \cite{dong2020benchmarking}, JPEG, and DiffPure \cite{nie2022diffusionmodelsadversarialpurification}).
    
    \item \textbf{Family 3: Multimodal Large Language Models.} We select 3 popular MLLMs: LLaVA-1.5-7B~\cite{liu2023llava}, Qwen-2.5-VL-7B~\cite{bai2025qwen25vltechnicalreport}, and Llama-3.2-11B-Vision\footnote{\url{https://ai.meta.com/blog/llama-3-2-connect-2024-vision-edge-mobile-devices/}}. These are used to further \textit{validate the generality} of the laws across different data modalities and architectures.
\end{itemize}

\subsubsection{Dataset and Evaluation Metrics}
\textbf{Dataset and Generation Protocol.}
Following previous work, we leverage the 1000 images from the ImageNet-1K NIPS 2017 dataset \cite{nips-2017-non-targeted-adversarial-attack, yang2022boostingtransferabilitytargetedadversarial}. To ensure the generality of our findings and create a large-scale testbed, we target 8 distinct categories for each source image \cite{zhang2020understandingadversarialexamplesmutual}. This creates a total of $1000 \times 8 = 8000$ unique source-target pairs for each ensemble configuration. As we generate adversarial examples for 7 distinct ensemble cardinalities $T \in \{2^0, 2^1, \dots, 2^6\}$, this results in a large-scale dataset of $7 \times 8000 = 56000$ generated adversarial examples.

To ensure our results are statistically robust and not biased by a single `lucky' or `unlucky' random selection of surrogate models, we employ a batched generation protocol. For any given cardinality $T$, we first partition the 1000 source images into 8 disjoint batches of 125 images. Then, for each batch, we randomly sample a \textit{new} ensemble of $T$ models from our surrogate pool. This newly sampled ensemble is used to craft the 125 adversarial examples (against all 8 target classes) for its assigned batch. This methodology ensures that the final metrics for each $T$ are averaged over 8 distinct, randomly-drawn ensembles, providing a highly robust and generalizable measurement of the scaling effect.

\textbf{Metrics.} For Family 1 and 2, we measure the Attack Success Rate (ASR), which is defined by the ratio between number of successful attacks (misclassified as target label) and total number of attacks. We also record the cross entropy loss between classifiers' predictions and target labels for Family 1 to demonstrate the validity of using ASR as the measurements of transferability. For Family 3 (MLLMs), which output text, we adopt a simple evaluation paradigm by string matching the target label text (e.g., `sea lion' for target label 150) and the MLLMs' descriptions. Since MLLMs produce descriptive text rather than precise classification results, we therefore adopt a relaxed keyword map (provided in Appendix~\ref{app:prompts}) to more robustly identify any semantic mentions of the target concept within the outputs.

\subsection{Log-Linear Scaling Laws in Image Classifiers}
\label{subsec:discovery}

\subsubsection{Empirical Observation}

We begin our analysis by evaluating the generated adversarial examples on the standard image classifiers. As demonstrated in Fig.~\ref{fig:scaling_laws} (top row), we identify a robust empirical phenomenon: the attack success rate (ASR) exhibits strong \textit{log-linear scaling laws} with the ensemble cardinality $T$.

As shown by the fitted lines, the ASR increases linearly as we increase $T$ on a logarithmic scale. Concurrently, the average cross-entropy loss against the target label (top row of Fig.~\ref{fig:scaling_laws}) exhibits a mirrored log-linear decay. This strong symmetric relationship confirms that ASR is a valid and robust proxy for the true optimization objective, as the decreasing loss directly correlates with the increasing success rate. This suggests that the effectiveness of ensemble attacks scales in a highly predictable manner, where each doubling of $T$ yields a consistent improvement in ASR.

\subsubsection{Analysis of the Log-Linear Relationship}

We empirically analyze the observable characteristics of these log-linear scaling laws:

\begin{itemize}
    \item \textbf{The Intercept (Baseline Vulnerability):} The starting point of the line (at $T=1$) represents the model's baseline vulnerability to single-model transfer attacks. As seen in Fig.~\ref{fig:scaling_laws}, this intercept is influenced by various factors, including the target model's architecture and the specific adversarial target.
    
    \item \textbf{The Slope (Scaling Efficiency):} The slope of the fitted line represents the `scaling efficiency' of the attack. It quantifies how much ASR is gained for each \textit{doubling} of $T$. Our results suggest this slope is a key characteristic of the target model's robustness to scaled attacks.
    
    \item \textbf{Domain of Validity:} Our observed scaling laws hold true within a specific domain. When $T$ is small ($2^0, 2^1$), the measured ASR shows large variance depending on the specific models randomly selected. The stable log-linear law emerges as a statistical result as $T$ increases. Meanwhile, as ASR approaches 1.0, the metric becomes saturated, and the linear trend naturally flattens. Our observations hold when ASR is below this saturation threshold.
\end{itemize}

\subsection{Verification of the Scaling Laws' Universality}
\label{subsec:verification}

\begin{figure}[t]
    \centering
    \begin{subfigure}[b]{0.48\columnwidth}
        \centering
        \includegraphics[width=\linewidth]{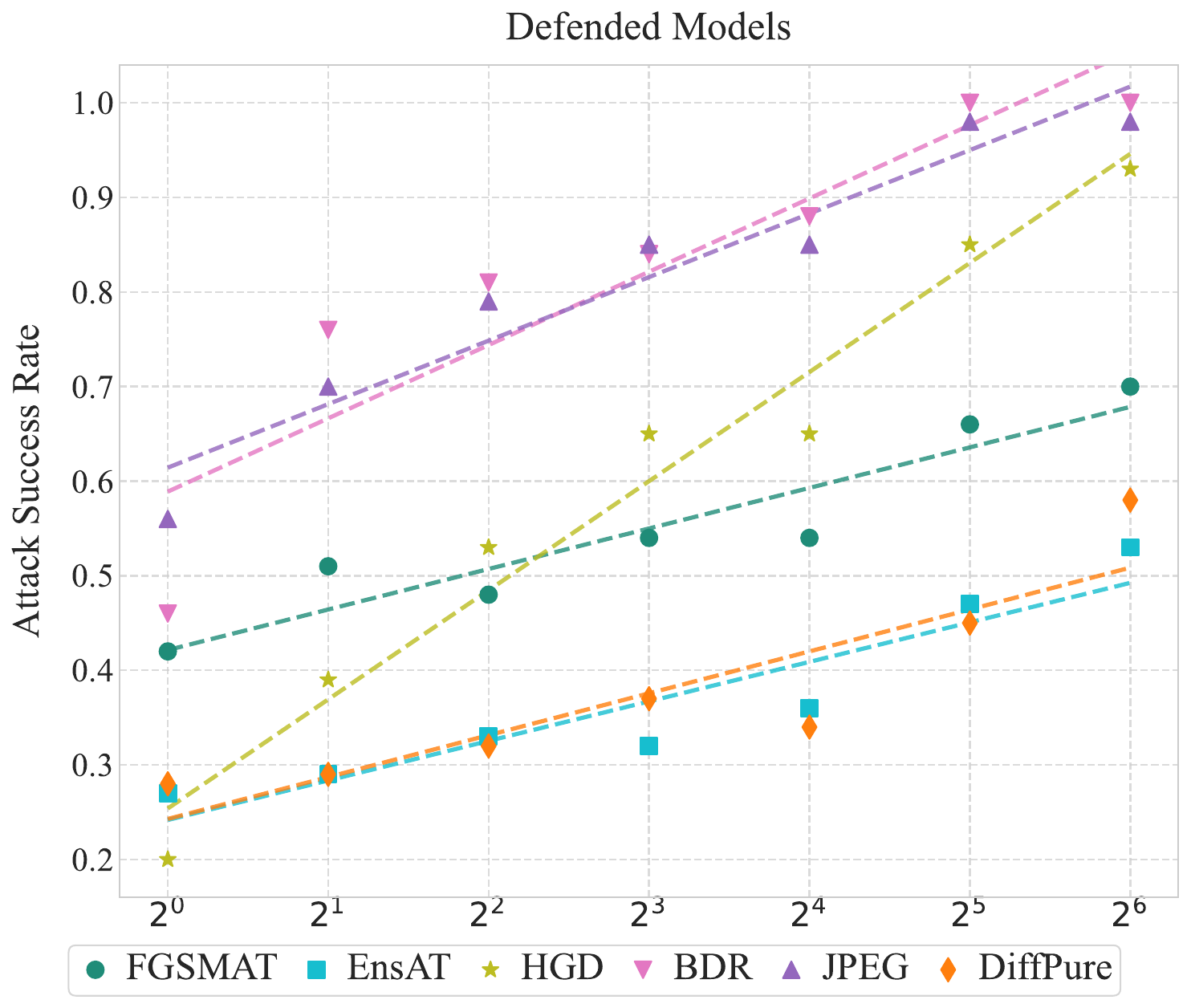}
        \caption{Scaling laws on Family 2.}
        \label{fig:defense_scaling_new}
    \end{subfigure}
    \hfill 
    \begin{subfigure}[b]{0.48\columnwidth}
        \centering
        \includegraphics[width=\linewidth]{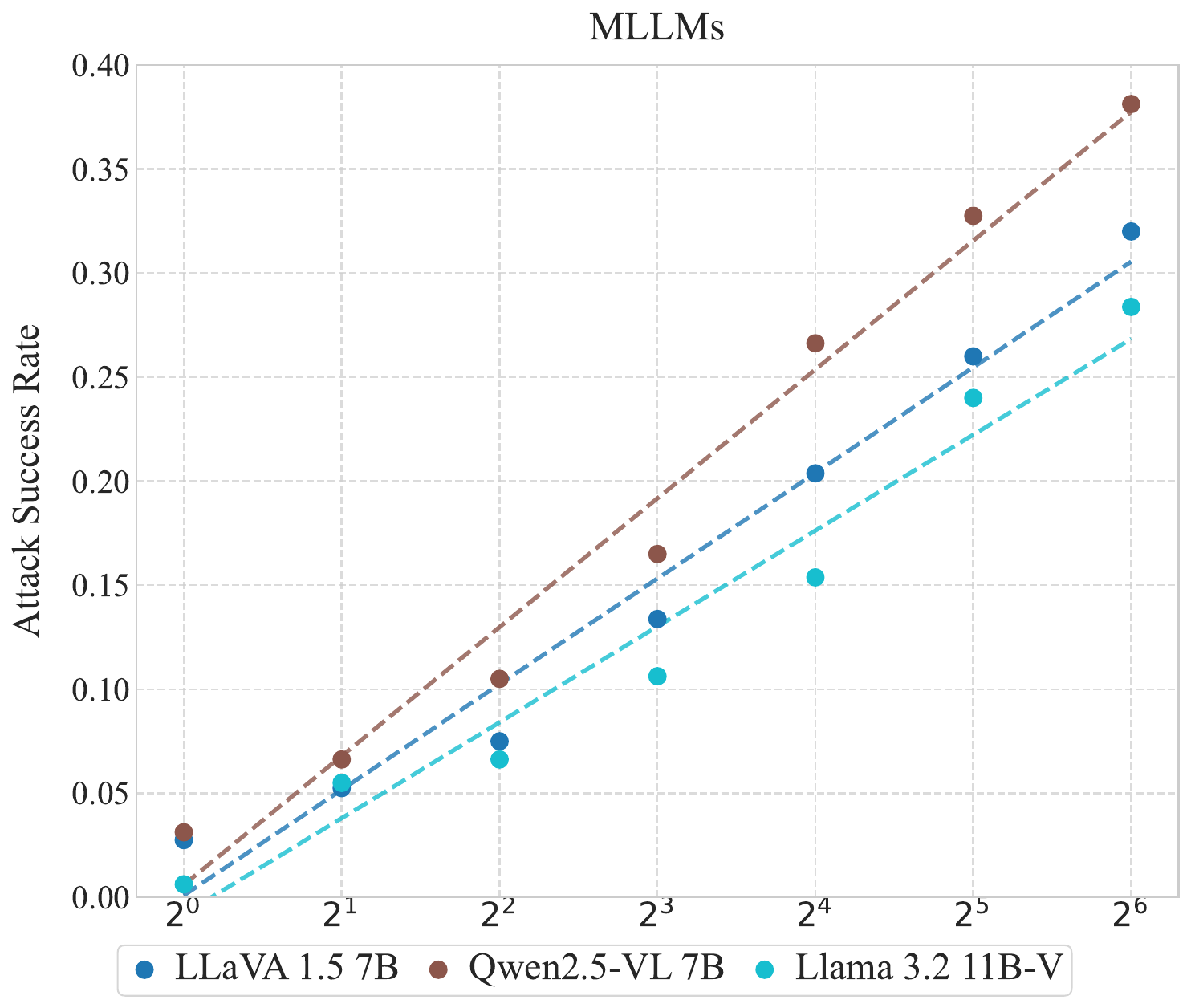}
        \caption{Scaling laws on Family 3.}
        \label{fig:vlm_scaling_new}
    \end{subfigure}
    \caption{Verification of the log-linear scaling laws' universality. (a) The trend persists against SOTA defenses. (b) The trend persists for MLLMs. In both settings, we record the ASR averaged across all 8 target classes to show overall model robustness.}
    \label{fig:verification_plots} 
    \vspace{-3ex}
\end{figure}

Having established the existence of these log-linear scaling laws, we now test their universality. A critical question is whether this trend is a fragile artifact of standard classifiers or a more fundamental principle of transfer attacks. We test this by evaluating their robustness against active defenses (Family 2) and their generality across entirely different architectures and modalities (Family 3).

\subsubsection{Robustness Against Defended Models}
\label{sec:defended}

We first test the adversarial examples against the defended models (Family 2). The key question is whether SOTA defenses can `break' this observed scaling trend.

As demonstrated in Fig.~\ref{fig:defense_scaling_new}, the log-linear scaling laws remain robustly intact. While some defenses successfully lower the overall ASR (seen in the lower ASR values across all $T$), they \textit{do not} eliminate the fundamental, predictable relationship between $\log T$ and ASR. 

This is a critical insight: defenses do not break the underlying scaling behavior. Instead, they primarily increase the `cost' of a successful attack (i.e., requiring a larger $T$ to reach the same ASR). This finding demonstrates that scaling the ensemble is a viable path to breaking even SOTA defenses. By leveraging these scaling laws, we improved the ASR of DiffPure~\cite{nie2022diffusionmodelsadversarialpurification} by over 30\%, adversarially trained models by over 20\%, and HGD~\cite{liao2018defenseadversarialattacksusing} by over 50\%.

\subsubsection{Generalization Across Architectures}
\label{sec:general_architectures}

Finally, we perform the ultimate test of generality: do the adversarial examples, generated \textit{only} from standard image classifiers, follow the same scaling laws on MLLMs?

Remarkably, as shown in Fig.~\ref{fig:vlm_scaling_new}, the log-linear scaling laws persist even in these advanced multimodal models. This cross-modal generalization is non-trivial, as the MLLMs possess vastly different architectures and are trained on different data modalities. This result strongly supports the generality of our scaling laws.

While the log-linear form is general, the specific robustness of each model varies. For instance, \texttt{LLaVA-1.5-7B} and \texttt{Qwen-2.5-VL-7B} have similar parameter counts ($\approx 7$B) and both show high robustness to single-model attacks, yet their averaged ASR curves clearly diverge, indicating different vulnerabilities to scaling. This suggests that the relationship between model scale and adversarial robustness is more complex in MLLMs and challenges the simple `model scale = robustness' narrative~\cite{bartoldson2024adversarialrobustnesslimitsscalinglaw}.

\section{Extending Scaling Laws to Vision Encoders}
\label{sec:vision_encoder}

\begin{figure*}[!t]
    \centering
    \includegraphics[width=\textwidth, trim=0cm 0cm 0cm 0cm, clip]{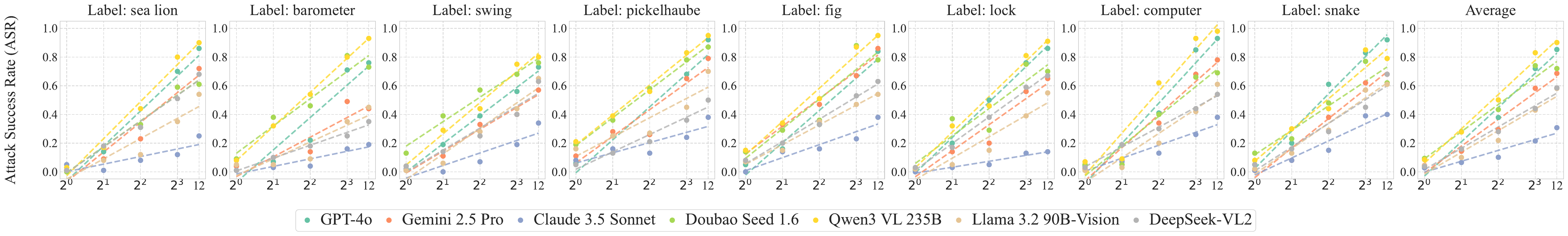}
    \caption{Verification of the log-linear scaling laws on SOTA MLLMs. We use an ensemble of CLIP models as surrogates, scaling the cardinality $T$ from 1 to 12. All attacks in this figure are generated using practical budget ($\epsilon = 16/255$). The ASR is measured using our single, strict LLM-as-Judge metric (defined in \cref{subsec:5.1}). The log-linear scaling laws clearly persist.}
    \vspace{-3ex}
    \label{fig:clip_LLM}
\end{figure*}

In this section, we test the generality and power of our discovered scaling laws. Having established the log-linear scaling laws on standard classifiers, we now extend our investigation to a more complex and practical domain: attacking modern MLLMs by targeting their foundational vision encoders, such as CLIP \cite{radford2021learningtransferablevisualmodels}. We hypothesize that the same scaling principles hold when using the embedding-space attack formulation defined in \cref{sec:clip_formulation}.

\subsection{Experimental Setup}
\label{subsec:5.1}

We first introduce our settings for attacking pretrained CLIP models and evaluating them on MLLM targets.

\textbf{Surrogate Models.}
We select 12 pretrained open-source CLIP models via OpenCLIP \cite{cherti2023reproducible}. In contrast to previous works that typically use fewer than 4 surrogate models~\cite{zhao2023evaluatingadversarialrobustnesslarge}, our scaled ensemble (up to $T=12$) is sufficient to yield robust results. The full model list is in Appendix~\ref{app:results_mllms}.

\textbf{Target Models.}
We select 7 state-of-the-art MLLMs as our black-box targets, including 4 commercial models (GPT-4o~\cite{openai2024gpt4technicalreport}, Claude-3.5-Sonnet\footnote{\url{https://www.anthropic.com/news/claude-3-family}}, Gemini-2.5-Pro~\cite{comanici2025gemini}, Doubao-Seed-1.6\footnote{\url{https://seed.bytedance.com/en/blog/introduction-to-techniques-used-in-seed1-6}}) and 3 state-of-the-art open-source models (Llama-3.2-90B-Vision-Instruct\footnote{\url{https://ai.meta.com/blog/llama-3-2-connect-2024-vision-edge-mobile-devices/}}, Qwen3-VL-235B-A22B-Instruct\footnote{\url{https://qwen.ai/blog?id=99f0335c4ad9ff6153e517418d48535ab6d8afef&from=research.latest-advancements-list}}, and DeepSeek-VL2~\cite{wu2024deepseekvl2mixtureofexpertsvisionlanguagemodels}).

\textbf{Dataset and Hyperparameters.}
We use the same SSA-CWA attack protocol as in \cref{subsec:setup}, but with a simplified dataset for this section. We randomly sample 100 images from the NIPS17 dataset with the same 8 distinct target classes, resulting in 800 unique source-target pairs per ensemble configuration. All experiments in this section are conducted using a unified perturbation budget of $\epsilon = 16/255$ under the $\ell_\infty$ norm, a common and widely applied setting for practical attacks \cite{dong2023robust, luo2024imageworth1000lies}.

\textbf{Evaluation Metric.}
We employ a single, unified evaluation metric for all MLLM targets. As MLLMs produce generative outputs instead of classification logits, we adopt the LLM-as-Judge framework from MultiTrust~\cite{zhang2024benchmarking}. We leverage GPT-4o to semantically evaluate the target model's description. An attack is considered successful only if GPT-4 judges that (1) the model's description mentions the adversarial target concept and (2) the description does not mention factors indicating the image is of low quality, noisy, or modified. The prompt of this strict, high-quality metric is provided in Appendix~\ref{app:prompts}.

\subsection{Generality and Power of Scaling Laws}
\label{subsec:5.2}

Our results, conducted on the latest generation of MLLMs, deliver two key findings.

\textbf{The log-linear scaling laws persist.} Fig.~\ref{fig:clip_LLM} presents the ASR results on our 7 SOTA MLLM targets as a function of ensemble cardinality ($T$). Our first key finding is that the log-linear scaling laws, discovered in Sec.~\ref{subsec:discovery}, robustly generalize to this new domain. This outcome confirms that the principle is not an artifact of standard classifiers but a fundamental property of ensemble attacks. Even when attacking the complex, SOTA vision encoders of models, scaling the number of CLIP surrogates ($T$) yields a predictable, linear increase in ASR on a log scale.

\textbf{Scaling is a powerful and practical weapon.}
Having verified the scaling laws (Fig.~\ref{fig:clip_LLM}), we now examine its practical impact. Table~\ref{tab:sota_mllm_results} reports the final ASR from our full scaled ensemble ($T=12$). The results are twofold: they confirm the devastating power of scaling while simultaneously revealing a stark hierarchy in SOTA model robustness. The attack's power is most evident on Qwen3-VL-235B and GPT-4o, which succumb to a 90.0\% and 85.0\% average ASR, respectively. This performance represents a significant leap in practical capability, surpassing the 40-50\% ASRs reported by previous methods on commercial models \cite{dong2023robust}. However, this effectiveness is not universal. Claude-3.5-Sonnet exhibits exceptional robustness, neutralizing the attack to a mere 31.0\% ASR. This resilience stands in sharp contrast to its direct competitor, Gemini-2.5-Pro, which is significantly more vulnerable at 69.0\% ASR. The open-source landscape mirrors this variability, with DeepSeek-VL2 (59.0\%) and Llama-3.2-90B (58.0\%) settling into a moderate robustness tier. These results demonstrate that scaling is a powerful and practical tool, and the resulting ASRs serve as a clear benchmark for the diverging robustness of modern MLLMs.

\newcolumntype{C}{>{\centering\arraybackslash}p{1.7cm}}

\begin{table*}[t]
\centering
\resizebox{\textwidth}{!}{
\begin{tabular}{l | C | C | C | C | C | C | C | C || c}
\toprule[1.5pt]
\textbf{Target Model - Label} & \textbf{Sea Lion} & \textbf{Barometer} & \textbf{Swing} & \textbf{Pickelhaube} & \textbf{Fig} & \textbf{Lock} & \textbf{Computer} & \textbf{Snake} & \textbf{Average} \\
\midrule
\quad GPT-4o & 0.86& 0.76& 0.73& 0.92& 0.84& 0.86& 0.93& 0.92& \textbf{0.85} \\
\quad Claude-3.5-Sonnet & 0.25& 0.19& 0.34& 0.38& 0.38& 0.14& 0.38& 0.41& \textbf{0.31} \\
\quad Gemini-2.5-Pro & 0.72& 0.44& 0.57& 0.79& 0.86& 0.65& 0.78& 0.68& \textbf{0.69} \\
\quad Doubao-Seed-1.6 & 0.61& 0.73& 0.76& 0.87& 0.78& 0.70& 0.69& 0.62& \textbf{0.72} \\
\midrule
\quad Llama-3.2-90B-Vision & 0.54& 0.45& 0.65& 0.70& 0.54& 0.55& 0.61& 0.61& \textbf{0.58} \\
\quad Qwen3-VL-235B & 0.90& 0.93& 0.80& 0.95& 0.95& 0.91& 0.98& 0.79& \textbf{0.90} \\
\quad DeepSeek-VL2 & 0.68& 0.35& 0.63& 0.50& 0.63& 0.67& 0.54& 0.68& \textbf{0.59} \\
\bottomrule[1.5pt]
\end{tabular}
}
\caption{Attack Success Rate (ASR) on SOTA MLLMs using our full scaled ensemble ($T=12$) and practical budget ($\epsilon = 16/255$). The ASR is measured using our unified LLM-as-Judge metric. We report the per-target ASR and the average ASR across all 8 target classes.}
\label{tab:sota_mllm_results}
\vspace{-3ex}
\end{table*}

\section{Further Analysis and Discussion}
\label{sec:discussion}

In this section, we further demonstrate the robustness of our discovered scaling laws through ablation studies (Sec.~\ref{sec:discussion:scaling_general}) and provide a deeper analysis into the semantic nature of the perturbations generated via scaling (Sec.~\ref{subsec:6.1}).

\subsection{Ablations on the Scaling Laws' Generality}
\label{sec:discussion:scaling_general}

While our main experiments (Sec.~\ref{sec:scaling}) focused on targeted attacks using SSA-CWA, here we verify that the \textit{log-linear scaling laws} are general principles that hold under different attack settings and across various optimizers.

\begin{figure}[h]
    \centering
    \begin{subfigure}{0.485\columnwidth}
        \centering
        \includegraphics[width=\linewidth]{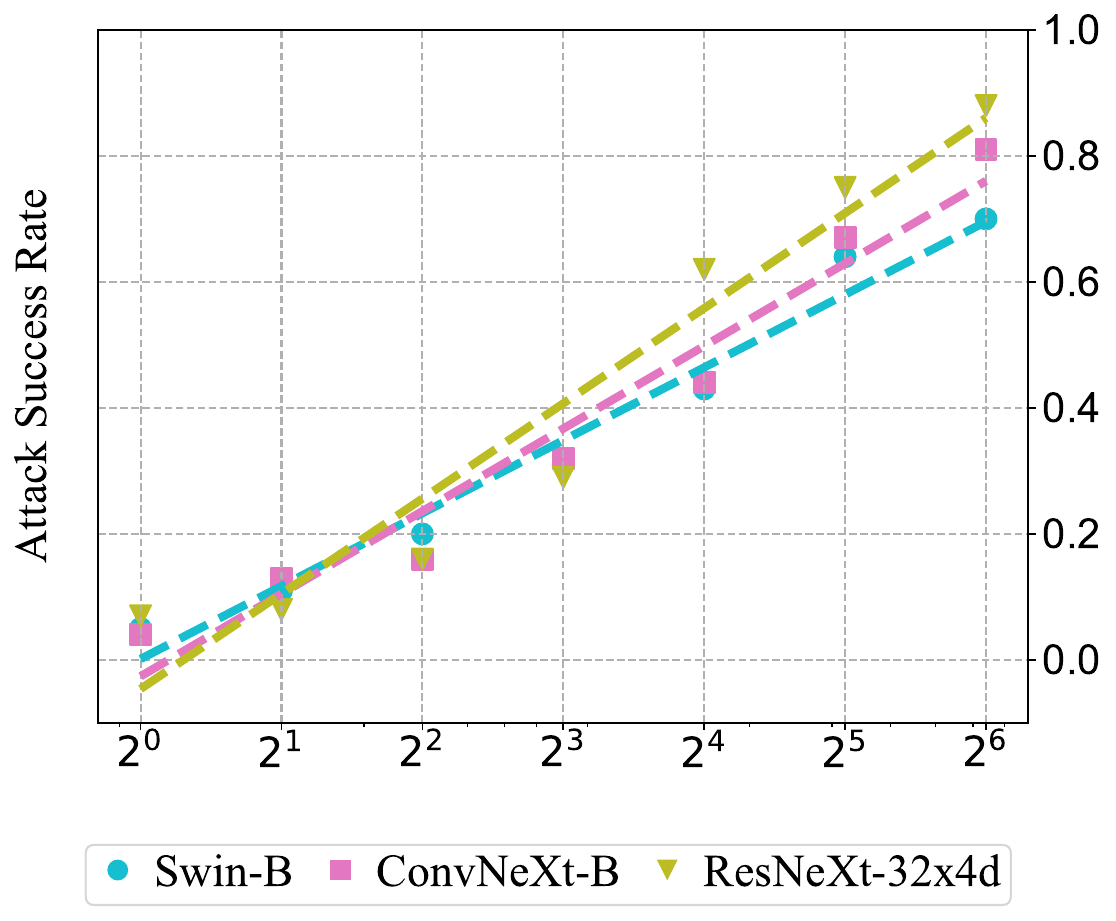}
        \caption{Under untargeted setting.}
        \label{fig:untargeted}
    \end{subfigure}
    \begin{subfigure}{0.475\columnwidth}
        \centering
        \includegraphics[width=\linewidth]{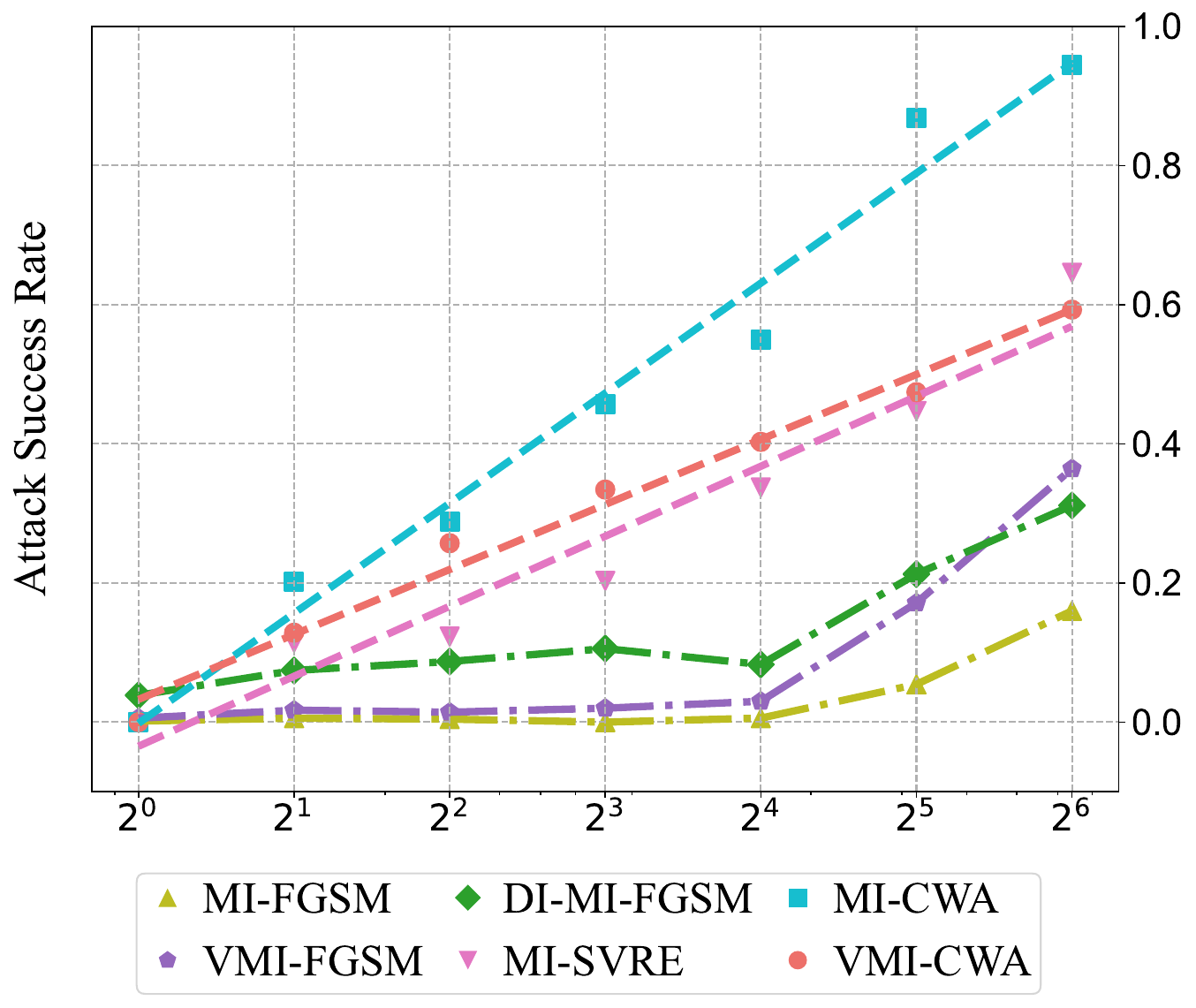}
        \caption{Across different optimizers.}
        \label{fig:algorithms_ablation}
    \end{subfigure}
    \caption{Ablations on the generality of the log-linear scaling laws. (a) The laws persist in the untargeted setting (Eq.~\ref{eq:empirical_loss_untargeted}). (b) The laws hold for advanced optimizers (MI-CWA, VMI-CWA, MI-SVRE) that mitigate gradient conflict. In contrast, the na\"ive logit ensemble strategy fails to scale, supporting our hypothesis in Sec.~\ref{sec:algorithms}.}
    \vspace{-3ex}
    \label{fig:ablations} 
\end{figure}

\subsubsection{Untargeted Attack Setting.}
We first validate the scaling laws in the untargeted setting (defined by Eq.~\ref{eq:empirical_loss_untargeted}). To do this, we follow the core experimental settings from \cref{subsec:setup}, using the SSA-CWA algorithm on the same 64-model surrogate pool and NIPS17 dataset. We measure the average ASR against 3 representative target models from Family 1 (Swin-B, ConvNeXt-B, and ResNeXt50). As the untargeted setting is a significantly simpler task, We use a small $\epsilon = 2 / 255$ to avoid the ASR quickly saturating at 100\%, which would obscure the underlying trend. As shown in Fig.~\ref{fig:untargeted}, even under untargeted conditions, a clear log-linear trend persists. This confirms that the scaling laws are a general principle of ensemble attacks, independent of the specific attack objective.

\subsubsection{Across Different Attack Algorithms.}
We now provide a tailored experiment to validate our hypothesis from \cref{sec:algorithms}: that the log-linear scaling laws are unlocked only by advanced ensemble optimizers (like CWA or SVRE) that resolve gradient conflict, while na\"ive ensemble strategies fail to scale.

\textbf{Experimental Design.}
To test this, we design two families of attack algorithms and evaluate their scaling performance. All experiments follow the protocol in \cref{subsec:setup} (targeting \texttt{sea lion}) and results are averaged over our standard target classifiers (Family 1 of Sec.~\ref{subsec:setup}).
\begin{itemize}
    \item Family 1 (Logit Ensemble Strategies): We test the logit averaging strategy, which we hypothesize will fail to scale. We combine the strategy with various powerful gradient optimizers and input transformations: MI-FGSM~\cite{dong2018boosting}, DI-MI-FGSM~\cite{xie2019improving}, and VMI-FGSM~\cite{wang2021enhancing}.
    \item Family 2 (Advanced Ensemble Strategies): We then test advanced gradient optimizers. This family includes MI-CWA~\cite{chen2023rethinking}, VMI-CWA~\cite{wang2021enhancing, chen2023rethinking}, and MI-SVRE~\cite{xiong2022stochastic}.
\end{itemize}

\textbf{Results and Analysis.}
The results are shown in Fig.~\ref{fig:algorithms_ablation}. All algorithms in Family 1, which use logit averaging, fail to exhibit the log-linear scaling law. While performance can be boosted by components like DI or VMI, their ASR stagnates at a low value. Conversely, algorithms in Family 2 demonstrate a clear and persistent log-linear scaling law. This behavior mirrors the results of our main SSA-CWA algorithm (from \cref{sec:scaling}), confirming that the scaling phenomenon is robust across different advanced optimizers, so long as they resolve the conflict. This experiment confirms our claim: the ability to scale an ensemble attack is contingent on the ensemble optimization strategy.

\subsection{The Semantic Nature of Scaled Perturbations}
\label{subsec:6.1}

\begin{figure}[t]
    \centering
    \includegraphics[width=\linewidth]{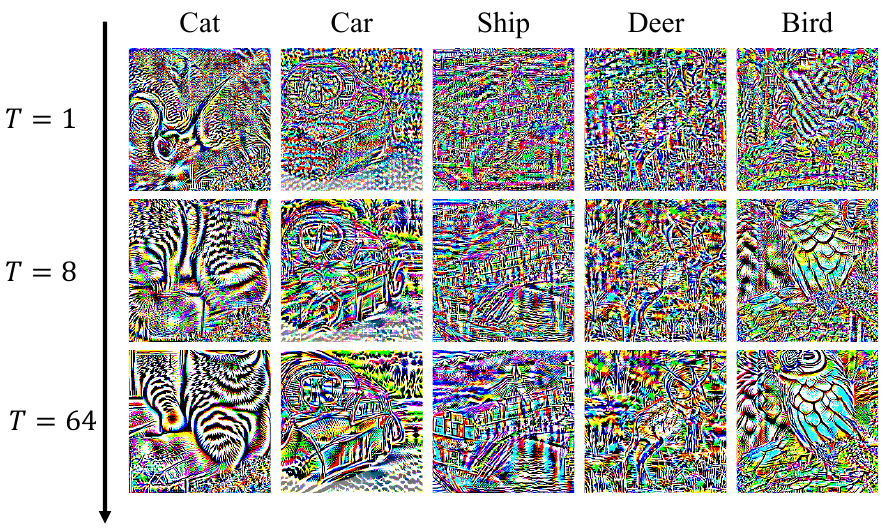}
    \caption{Visualization of adversarial perturbations as $T$ scales. We take 5 labels from CIFAR10~\cite{krizhevsky2009learning} to create adversarial examples as they are relatively easy for human eyes to recognize. As the ensemble cardinality grows (top to bottom), the perturbations evolve from unstructured noise ($T=1$) into semantically meaningful patterns that resemble features of target classes ($T=64$).}
    \label{fig:visualizations}
    \vspace{-3ex}
\end{figure}

Beyond attack success rates, we investigate \textit{what} the ensemble attack learns as $T$ increases. As shown in Fig.~\ref{fig:visualizations}, scaling the ensemble fundamentally changes the \textit{nature} of the resulting perturbation. At $T=1$, the perturbation is noisy and model-specific. However, as $T$ increases to 64, the perturbation evolves to contain clear, interpretable, and semantic features of the target class (e.g., textures for `Cat', an owl-like shape for `Bird').

This is a critical finding that provides a qualitative explanation for the scaling laws. First, scaling $T$ forces the optimizer to discard model-specific, `non-robust features' \cite{ilyas2019adversarialexamplesbugsfeatures} and instead find a `common weakness' solution that fools all models. Next, our visualization strongly suggests this common weakness is not just a mathematical artifact but is tied to the \textit{semantic essence} of the target class. The attack learns to inject robust features of the target \cite{dai2022fundamental}.

This reinforces that our scaling approach is a principled method to distill the shared, transferable features of a concept from a diverse set of models. As noted by \cite{fawzi2018adversarial}, images lie on a low-dimensional manifold; our scaling approach appears to provide a more accurate and robust representation of this manifold for the target class.

\section{Conclusion}
\label{sec:conclusion}

In this work, we investigate scaling laws for black-box adversarial attacks. We find that while gradient conflict stalls na\"ive ensembling, advanced optimizers (e.g., CWA) unlock a robust log-linear scaling law: the Attack Success Rate increases linearly with $\log T$. We empirically verify this law's universality across standard classifiers (Sec.~\ref{subsec:discovery}), SOTA defenses , and MLLMs (Sec.~\ref{subsec:verification}), and show that scaling distills semantic features (Sec.~\ref{subsec:6.1}). Applying this methodology to SOTA MLLMs (Sec.~\ref{sec:vision_encoder}) serves as a powerful benchmark. The scaled attack achieves 85.0\% ASR on GPT-4o, while Claude-3.5-Sonnet shows exceptional resilience at only 31.0\% ASR. Our findings urge a shift in focus for robustness evaluation: from designing intricate algorithms on small ensembles to understanding the principled and powerful threat of scaling, which is crucial for building and verifying genuinely robust foundation models.

{
    \small
    \bibliographystyle{ieeenat_fullname}
    \bibliography{main}
}

\clearpage
\appendix

\section{Proof of Theorem 1.}
\label{app:proof_thm1}

As defined in \cref{def:minimizer}, \(\bm{x}^*\) is the minimizer of the population loss \(\mathbb{L}(\bm{x})=\mathbb{E}_{f \in \mathcal{F}} [\mathcal{L} (f(\bm{x}), \bm{y})]\), and \(\hat{\bm{x}}\) is the minimizer of the empirical loss \(\hat{\mathbb{L}}(\bm{x})=\frac{1}{T} \sum_{i=1}^{T} \mathcal{L} (f_i(\bm{x}), \bm{y})\). Thereby we have by gradient equals 0:
\begin{equation*}
    \nabla \mathbb{L}(x^*) = \nabla \hat{\mathbb{L}}(\hat{\bm{x}}) = 0
\end{equation*}
We use Taylor expansion of \(\nabla \hat{\mathbb{L}}\) at $\hat{\bm{x}}$:
\begin{equation*}
    0=\nabla \hat{\mathbb{L}}(\hat{\bm{x}})=\nabla \hat{\mathbb{L}}(\bm{x}^*) + \nabla^2 \hat{\mathbb{L}}(\bm{x}^*) (\hat{\bm{x}}-\bm{x}^*) + \mathcal{O}(\|\hat{\bm{x}}-\bm{x}^*\|^2).
\end{equation*}
Rearranging, we have:
\begin{equation*}
    \sqrt{T}(\hat{\bm{x}}-\bm{x}^*) \approx - \sqrt{T}(\nabla^2 \hat{\mathbb{L}}(\bm{x}^*))^{-1}\nabla \hat{\mathbb{L}}(\bm{x}^*).
\end{equation*}
From Central Limit Theorem:
\begin{equation*}
   \begin{aligned}
       \sqrt{T}\nabla \hat{\mathbb{L}}(\bm{x}^*)&= \sqrt{T}(\nabla \hat{\mathbb{L}}(\bm{x}^*)-\nabla \mathbb{L}(\bm{x}^*)) \\ &\overset{d}{\to} \mathcal{N}(0, \mathrm{Cov}(\mathbb{L}(f_i(\bm{x}^*)))),
   \end{aligned}
\end{equation*}
and the law of large numbers:
\begin{equation*}
    \nabla^2 \hat{\mathbb{L}}(\bm{x}^*) \overset{p}{\to} \nabla^2 \mathbb{L}(\bm{x}^*).
\end{equation*}
Plug into the previous equation using \(A\mathcal{N}(0, \Sigma)=\mathcal{N}(0, A\Sigma A^T)\):
\begin{equation*}
    \begin{aligned}
        &\sqrt{T}(\hat{\bm{x}}-\bm{x}^*) \overset{d}{\to}(\nabla^2 \mathbb{L}(\bm{x}^*))^{-1}\mathcal{N}(0, \mathrm{Cov}(\mathbb{L}(f_i(\bm{x}^*)))) \\
        &\overset{d}{=} \mathcal{N}(0,\nabla^2 \mathbb{L}(\bm{x}^*)^{-1} \mathrm{Cov}(\mathbb{L}(f_i(\bm{x}^*)) \nabla^2 \mathbb{L}(\bm{x}^*)^{-1}) .
    \end{aligned}
\end{equation*}
The above indicate \(\hat{\bm{x}}-\bm{x}^*\) converges in \(\mathcal{O}(\frac{1}{\sqrt{T}})\). \\
We can also use Taylor expansion of $\mathbb{L}$ at \(\bm{x}^*\):
\begin{equation*}
    \mathbb{L}(\hat{\bm{x}})\approx\mathbb{L}(\bm{x}^*) + 0 + \frac{1}{2}(\hat{\bm{x}} - \bm{x}^*)^T\nabla^2 \mathbb{L}(\bm{x}^*)(\hat{\bm{x}} - \bm{x}^*).
\end{equation*}
Rearranging, we have:
\begin{equation*}
    \begin{aligned}
        T(\mathbb{L}(\hat{\bm{x}})- \mathbb{L}(\bm{x}^*) )&\approx \frac{1}{2}\sqrt{T}(\hat{\bm{x}} - \bm{x}^*)^T\nabla^2 \mathbb{L}(\bm{x}^*)\sqrt{T}(\hat{\bm{x}} - \bm{x}^*) \\
        &=\frac{1}{2} \|\nabla^2 \mathbb{L}(\bm{x}^*)\sqrt{T}(\hat{\bm{x}} - \bm{x}^*) \|_2^2=\frac{1}{2} \|S\|_2^2,
    \end{aligned}
\end{equation*}
where
\begin{equation*}
    S \sim \mathcal{N}(0,\nabla^2 \mathbb{L}(\bm{x}^*)^{-1/2} \mathrm{Cov}(\mathbb{L}(f_i(\bm{x}^*)) \nabla^2 \mathbb{L}(\bm{x}^*)^{-1/2}).
\end{equation*}
Thus \( T(\mathbb{L}(\hat{\bm{x}})- \mathbb{L}(\bm{x}^*) )\) converges into a chi-square distribution, which means \(\mathbb{L}(\hat{\bm{x}})- \mathbb{L}(\bm{x}^*) \) converges in \(\mathcal{O}(\frac{1}{T})\).

\section{Experiment Settings for Scaling Laws}
\label{app:settings_scaling_laws}

In this section we list our models that are included in Sec.~\ref{subsec:setup}. We select these models based on their diversity in architectures, training methods, and training datasets. Results are shown in Table~\ref{tab:subrows} and Table~\ref{tab:subrows_2}, where all of our models come from open-source platforms such as Torchvision~\cite{torchvision2016} and Timm~\cite{rw2019timm}. Note that \textit{default} in the table means normally trained on ImageNet-1K dataset.

\begin{table}[h]
\centering
\resizebox{\columnwidth}{!}{ 
\begin{tabular}{c | c | c}
\toprule[1.5pt]
Model name & Architecture & Settings\\
\midrule
AlexNet & alexnet & default\\
\midrule
\multirow{3}{*}{DenseNet} & densenet121 &default\\ 
                         & densenet161 &default\\
                         & densenet169 &default\\
\midrule
\multirow{6}{*}{EfficientNet} & efficientnet-b0 &default\\ 
                         & efficientnet-b1 &default\\
                         & efficientnet-b4 &default\\
                         & efficientnet-v2-s &default\\
                         & efficientnet-v2-m &default\\
                         & efficientnet-v2-l &default\\
\midrule
GoogleNet & googlenet &default\\
\midrule
Inception & inception-v3 &default\\
\midrule
MaxVit & maxvit-t &default\\
\midrule
\multirow{4}{*}{MnasNet} & mnasnet-0-5 &default\\
                        & mnasnet-0-75 &default\\ 
                        & mnasnet-1-0 &default\\
                         & mnasnet-1-3 &default\\ 
\midrule
\multirow{3}{*}{MobileNet} & mobilenet-v2 &default\\
                        & mobilenet-v3-s &default\\ 
                         & mobilenet-v3-l &default\\ 
\midrule
\multirow{8}{*}{RegNet} & regnet-x-8gf &default\\
    & regnet-x-32gf &default\\
    & regnet-x-400mf &default\\ 
    & regnet-x-800mf &default\\
    & regnet-y-16gf &default\\
    & regnet-y-32gf &default\\
    & regnet-y-400mf &default\\
    & regnet-y-800mf &default\\
\midrule
\multirow{5}{*}{ResNet} & resnet18 &default\\
                        & resnet34 &default\\
                        & resnet50 &default\\ 
                         & resnet101 &default\\ 
                         & resnet152 &default\\
\midrule
\multirow{4}{*}{ShuffleNet} & shufflenet-v2-x0.5 &default\\
           & shufflenet-v2-x1.0 &default\\
            & shufflenet-v2-x1.5 &default\\
            & shufflenet-v2-x2.0 &default\\
\midrule
\multirow{2}{*}{SqueezeNet} & squeezenet-1.0 &default\\
                            & squeezenet-1.1 &default\\
\midrule
\multirow{4}{*}{ViT}    &  vit-b-16 & default \\ 
                         & vit-b-32  &default \\
                         & vit-l-16 &default\\
                         & vit-l-32 &default\\
                         & vit-b-16 & pretrained on ImageNet-21K\\
\bottomrule[1.5pt]
\end{tabular}}
\caption{List of pretrained image classifiers that are selected for our model ensemble.}
\label{tab:subrows}
\end{table}

\begin{table}[t]
\centering
\resizebox{\columnwidth}{!}{ 
\begin{tabular}{c|c|c}
\toprule[1.5pt]
Model name & Architecture & Settings\\
\midrule
\multirow{7}{*}{Vgg}  & vgg11 &default\\
                        & vgg11-bn &default\\
                        & vgg13 &default\\
                        & vgg13-bn &default\\ 
                         & vgg16 &default\\ 
                        & vgg16-bn &default\\
                         & vgg19 &default\\
\midrule
\multirow{2}{*}{Wide-ResNet} & wide-resnet50-2 &default\\ 
                         & wide-resnet101-2 &default\\
\midrule
BeiT & beit-B-16 & pretrained on ImageNet-21k\\
\midrule
GhostNet & ghostnet-100& default\\
\midrule
DeiT & deit-B-16 &default\\
\midrule
LCnet & lcnet-050 &default\\
\midrule
RepVgg & repvgg-a2& default\\
\midrule
DPN & dpn98& default\\
\midrule
SegFormer & mit-b0 &default\\
\midrule
BiT & bit-50& default\\
\midrule
CvT & cvt-13 &default\\
\midrule
Twins-SVT & twins-svt-large & default \\
\midrule
NFNet & nfnet-l0 & default \\
\bottomrule[1.5pt]
\end{tabular}}
\caption{List of pretrained image classifiers that are selected for our model ensemble (continued).}
\label{tab:subrows_2}
\end{table}

\section{SSA-CWA Algorithm Declaration}
\label{app:ssa_cwa}

We provide the pseudo-code version of SSA-CWA implementation in Algorithm~\ref{algo:SSACWA}. Their combination is induced from ~\cite{long2022frequency} and ~\cite{chen2023rethinking}, while implementations of other algorithms (e.g., MI-FGSM, MI-CWA) can be found in their corresponding papers.

\begin{algorithm}
\caption{SSA-CWA Algorithm}
\label{algo:SSACWA}
\begin{algorithmic}[1]
\State \textbf{Input:} natural image $\bm{x}_{\text{nat}}$, label $\bm{y}$, $L_{\infty}$ constraint $\epsilon$, iterations $T$, loss function $\mathcal{L}$, spectrum transformation function $\mathcal{T}$, model ensemble $\{f_i\}_{i=1}^N$, spectrum transformation times $M$,  decay factor $\mu$, step sizes $\alpha, \beta$. 
\State \textbf{Output:} An adversarial example \( \bm{x} \)
\For{t = 0 \textbf{to} T-1}
    \State Update $\bm{x_t}$ by $\bm{x}^0_t = x_{t-1}$
    \For{i = 1 \textbf{to} N}
        \For{j = 1 \textbf{to} M} \Comment{Spectrum transformation}
            \State Obtain transformed image \( \mathcal{T}(\bm{x}^{j}_{t,i}) \)
            \State Compute \( \bm{g}_i^j = \nabla_{\bm{x}^{j}_{t,i}} \mathcal{L}(\mathcal{T}(\bm{x}^{j}_{t,i}), \bm{y}) \)
        \EndFor
        \State Compute gradient $\bm{g}_i = \frac{1}{M} \sum_{j=1}^M \bm{g}^j_i$
        \State Update inner momentum by $\bm{m} = \mu \cdot \bm{m} + \frac{\bm{g}_i}{\left\lVert \bm{g}_i \right\rVert_2}$
        \State Update \( \bm{x}^i_t = \text{clip}(\bm{x}_{\text{nat}}, \epsilon) \{ \bm{x}^{i-1}_t + \beta \cdot \bm{m} \} \)
    \EndFor
    \State Compute gradient \( \bm{g} = \bm{x}_t^N - \bm{x}_t^0 \)
    \State Update outer momentum by $\bm{h} = \mu \cdot \bm{h} + \frac{\bm{g}}{\left\lVert \bm{g} \right\rVert_2}$
    \State $\bm{x}_t =\text{clip}(\bm{x}_{\text{nat}}, \epsilon) \{ \bm{x}_t^N + \alpha \cdot \text{sign}(\bm{h}) \}$
\EndFor
\State \textbf{Return:} \( \bm{x}_{T-1} \)
\end{algorithmic}
\end{algorithm}




\section{Evaluation Metrics Details}
\label{app:prompts}

As MLLMs output generative text rather than classification logits, assessing attack success requires parsing natural language descriptions. In this work, we employ two distinct evaluation paradigms tailored to the specific experimental goals in Sec.~\ref{sec:scaling} and Sec.~\ref{sec:vision_encoder}.

In Sec.~\ref{sec:general_architectures} (Family 3: MLLMs), we evaluate the scaling laws using a string-matching approach. Since MLLMs often produce descriptive text rather than the precise ImageNet class names, strict exact matching can lead to false negatives. Therefore, we adopt a relaxed keyword map. An attack is considered successful if the MLLM's output contains any of the keywords associated with the target label.

Table~\ref{tab:keyword_map} lists the target labels used in our experiments and their corresponding relaxed keywords.

\begin{table}[h]
    \centering
    \resizebox{\columnwidth}{!}{
    \begin{tabular}{l | l}
        \toprule
        \textbf{Target Label (ImageNet)} & \textbf{Accepted Keywords (Case-Insensitive)} \\
        \midrule
        Sea Lion & sea lion, seal, marine mammal \\
        Barometer & barometer, gauge, measuring instrument, clock \\
        Swing & swing, playground, seat \\
        Pickelhaube & pickelhaube, helmet, spiked helmet, military hat \\
        Fig & fig, fruit, plant \\
        Lock & lock, padlock, dial \\
        Computer & computer, pc, laptop, desktop, phone, device \\
        Snake & snake, serpent, reptile, python, cobra \\
        \bottomrule
    \end{tabular}
    }
    \caption{The relaxed keyword map used for evaluating MLLMs in Sec.~\ref{sec:scaling}. We map the specific ImageNet class label to a set of semantically equivalent or broader keywords.}
    \label{tab:keyword_map}
\end{table}

In Sec.~\ref{sec:vision_encoder}, to rigorously benchmark SOTA MLLMs, we adopt the LLM-as-Judge framework inspired by MultiTrust~\cite{zhang2024benchmarking}. We use GPT-4o as the judge to evaluate the quality and success of the attack. 

The evaluation is strict: an attack is successful \textit{only if} the description mentions the target and does not mention artifacts of the attack (noise, distortion). Fig.~\ref{fig:gpt4_prompt} presents the exact system prompt used in our experiments.

\begin{figure}[h]
    \centering
    \fbox{
    \begin{minipage}{0.95\columnwidth}
        \textbf{Prompt for GPT-4o Judge Correctness:}
        
        Given a chatbot's description of a scene: \texttt{[response]}. If it is stated in the description that there is a/an \texttt{[label]}? Please answer 'Yes.' or 'No.'.
        
        \vspace{0.2cm}
        \textbf{Prompt for GPT-4o Judge Image Quality:}
        
        Given a chatbot's description of an image: \texttt{[response]}. Does the description mention factors that affect the judgment of the image content, such as low image quality, noise, or modification? Please answer 'Yes.' or 'No.'.
    \end{minipage}
    }
    \caption{The GPT-4o system prompt used for the unified evaluation metric in Sec.~\ref{sec:vision_encoder}. This metric ensures that successful attacks are both effective (semantics changed) and stealthy (no obvious noise detected by the model).}
    \label{fig:gpt4_prompt}
\end{figure}

\section{Experiment Settings for Attacking MLLMs}
\label{app:results_mllms}

\begin{table}[h]
\centering
\resizebox{\columnwidth}{!}{ 
\begin{tabular}{c | c | c}
\toprule[1.5pt]
Model Name & Architecture & Datasets \\
\midrule
ViT-L OpenAI& ViT-L-14 & WebImageText \\
ViT-B LAION-2B & ViT-B-32 & LAION-2B \\
ViT-B LAION-2B& ViT-B-16 & LAION-2B \\
ViT-bigG LAION-2B& ViT-bigG-14 & LAION-2B \\
ViT-g LAION-2B & ViT-g-14 & LAION-2B \\
ViT-H LAION-2B& ViT-H-14 & LAION-2B \\
ViT-L LAION-2B & ViT-L-14 & LAION-2B \\
ConvNeXt-L  LAION-2B & ConvNeXt-large-d-320 & LAION-2B \\
ViT-B DataComp & ViT-B-16 & DataComp-1B \\
 ViT-B DataComp& ViT-B-32 & DataComp-1B \\
ViT-L DataComp& ViT-L-14 & DataComp-1B \\
SigLip WebLI & ViT-SO400M-14-SigLIP & WebLI \\
\bottomrule[1.5pt]
\end{tabular}}
\caption{List of CLIPs that are selected as model ensemble.}
\label{tab:12_clips}
\end{table}


In Table.~\ref{tab:12_clips} we list the 12 CLIPs that are collected for our model ensemble, which are all open-source models selected from OpenAI \cite{radford2021learningtransferablevisualmodels}, OpenCLIP \cite{cherti2023reproducible}, and Timm \cite{rw2019timm}.




\section{Related Work}
\label{2_relatedWork}

\subsection{Black-Box Attacks}

Deep learning models are widely recognized for their vulnerability to adversarial attacks \cite{szegedy2013intriguing, goodfellow2014explaining}. Extensive research has documented this susceptibility, particularly in image classification contexts \cite{athalye2018obfuscated, carlini2017towards}, under two principal settings: white-box and black-box attacks. In the white-box setting, the adversary has complete access to the target model \cite{madry2017towards}. Conversely, in the black-box setting, such information is obscured from the attacker. With limited access to the target model, black-box attacks typically either construct queries and rely on feedback from the model \cite{dong2021query, ilyas2018black} or leverage the transferability of predefined surrogate models \cite{liu2016delving}. Query-based attacks often suffer from high time complexity and substantial computational costs, making transfer-based methods more practical and cost-effective for generating adversarial examples, as they do not necessitate direct interaction with target models during the training phase.

Within the realm of transfer-based attacks, existing methods can generally be categorized into three types: input transformation, gradient-based optimization, and ensemble attacks. Input transformation methods employ data augmentation techniques before feeding inputs into models, such as resizing and padding \cite{xie2019improving}, translations \cite{dong2019evading}, or transforming them into frequency domain \cite{long2022frequency}. Gradient-based methods concentrate on designing optimization algorithms to improve transferability, incorporating strategies such as momentum optimization \cite{dong2018boosting, lin2019nesterov}, scheduled step sizes \cite{gao2020patch}, and gradient variance reduction \cite{wang2021enhancing}.

\subsection{Ensemble Attacks}

Inspired by model ensemble techniques in machine learning \cite{wortsman2022model, dietterich2000ensemble, gontijo2021no}, researchers have developed sophisticated adversarial attack strategies utilizing sets of surrogate models. In forming an ensemble paradigm, practitioners often average over losses \cite{dong2018boosting}, predictions \cite{liu2016delving}, or logits \cite{dong2018boosting}. Additionally, advanced ensemble algorithms have been proposed to enhance adversarial transferability \cite{li2023making, chen2023rethinking, xiong2022stochastic, chen2023adaptive}. However, these approaches tend to treat model ensemble as a mere technique for improving transferability without delving into the underlying principles.

Prior studies \cite{huang2023t} indicate that increasing the number of classifiers in adversarial attacks can effectively reduce the upper bound of generalization error in Empirical Risk Minimization. Recent research \cite{yao2025understandingmodelensembletransferable} has theoretically defined transferability error and sought to minimize it through model ensemble. Nonetheless, while an upper bound is given, they do not really scale up in practice and have to train neural networks from scratch to meet the assumption of sufficient diversity among ensemble models \cite{liu2019deep}. 

\subsection{Scaling Laws}

In recent years, the concept of scaling laws has gained significant attention. 
These laws describe the relationship between model size -- measured in terms of parameters, data, or training FLOPs -- and generalization ability, typically reflected in evaluation loss \cite{kaplan2020scalinglawsneurallanguage}. Subsequent studies have expanded on scaling laws from various perspectives, including image classification \cite{bahri2024explaining, hestness2017deep}, language modeling \cite{sharma2022scaling} and mixture of experts (MoE) models \cite{sun2024hunyuanlargeopensourcemoemodel}, highlighting the fundamental nature of this empirical relationship.

In the domain of adversarial attack research, recent work has attempted to derive scaling laws for adversarially trained defense models on CIFAR-10 \cite{krizhevsky2009learning, bartoldson2024adversarialrobustnesslimitsscalinglaw}. However, the broader concept of scaling remains underexplored. While some preliminary observations on the effect of enlarging model ensembles can be found in prior studies \cite{yao2025understandingmodelensembletransferable}, these effects have not been systematically analyzed. Recent work \citet{wei2023jailbreak} also explore the scaling law of attack success rate when jailbreaking large language models~\cite{zou2023universal,chen2025towards} by increasing number of in-context demonstrations 
In this paper, however, we focus on the empirical scaling laws within model ensembles, and verify it through extensive experiments on open-source pretrained models. Motivated by this perspective, we unveil vulnerabilities within commercial LLMs by targeting their vision encoders.

\end{document}